\documentclass{sigchi-ext}
\usepackage[T1]{fontenc}
\usepackage{textcomp}
\usepackage[scaled=.92]{helvet} 
\usepackage{graphicx} 
\usepackage{balance}  
\usepackage{booktabs} 
\usepackage{ccicons}  
\usepackage{ragged2e} 
\usepackage{hyperref}

\def\plainkeywords{Fairness and Bias in Artificial Intelligence; Ethics of AI}

\title{Explaining How Your AI System is Fair}

\numberofauthors{2}
\author{%
  \alignauthor{%
    \textbf{Boris Ruf}\\
    \affaddr{Research Data Scientist} \\
    \affaddr{AXA GO, AI Research} \\
    \affaddr{Paris, France} \\
    \email{boris.ruf@axa.com} } \vfil 
    \alignauthor{%
    \textbf{Marcin Detyniecki}\\
    \affaddr{Head of R\&D} \\
    \affaddr{AXA GO, AI Research}\\
    \affaddr{Paris, France}\\
    \email{marcin.detyniecki@axa.com} }}

\begin{document}

\CopyrightYear{2021}
\setcopyright{rightsretained}
\conferenceinfo{CHI'21,}{May  8--13, 2021, Online Virtual Conference (originally Yokohama, Japan)}
\isbn{978-1-4503-6819-3/20/04}
\doi{https://doi.org/10.1145/3334480.XXXXXXX}
\copyrightinfo{\acmcopyright}

\maketitle

\RaggedRight{} 


\begin{abstract}
To implement fair machine learning in a sustainable way, choosing the right fairness objective is key. Since fairness is a concept of justice which comes in various, sometimes conflicting definitions, this is not a trivial task though. The most appropriate fairness definition for an artificial intelligence (AI) system is a matter of ethical standards and legal requirements, and the right choice depends on the particular use case and its context.\\
In this position paper, we propose to use a decision tree as means to explain and justify the implemented kind of fairness to the end users. Such a structure would first of all support AI practitioners in mapping ethical principles to fairness definitions for a concrete application and therefore make the selection a straightforward and transparent process. However, this approach would also help document the reasoning behind the decision making. Due to the general complexity of the topic of fairness in AI, we argue that specifying "fairness" for a given use case is the best way forward to maintain confidence in AI systems. In this case, this could be achieved by sharing the reasons and principles expressed during the decision making process with the broader audience.

\end{abstract}

\keywords{\plainkeywords}

\ccsdesc[500]{Computing methodologies~Philosophical/theoretical foundations of artificial intelligence}

\printccsdesc

\section{Introduction}
AI systems have vast potential to enrich our lives in all sorts of ways, but they also introduce new and changed risks. One of them being the problem that AI systems can reproduce and even reinforce unwanted bias~\cite{DBLP:journals/corr/abs-1301-6822, DBLP:journals/corr/BolukbasiCZSK16a, DBLP:journals/corr/IslamBN16, 2017arXiv170309207B}. Plenty of mathematical definitions of fairness and many technical mitigation strategies have been proposed during the last years~\cite{mehrabi2019survey}. While those definitions reflect existing, widespread notions of fairness in society, they do encode quite different objectives, and satisfying all of them at the same time was shown to be impossible~\cite{Friedler2016,Corbett-Davies2018}. Thus, the question of how to select the best option is now the challenge. However, there is no universal answer, because the ideal fairness definition always depends on the use case and its context, and it is a matter of ethical criteria and legal requirements~\cite{Holstein2019}. Unfortunately, this insight has further complicated the task of reassuring end users that AI systems treat them in a fair way: Now, it is not only about explaining the characteristics of an implemented kind of fairness, but also about why this definition was considered to be the best one.



\section{Select the Right Fairness Definition}
AI stakeholders need to weigh several conflicting ethical principles against each other in order to make the right choice. Many decision points have been found to partition the complex landscape of fairness definitions, for example questions around applicable policies, the presence of ground truth data, the risk of label bias or the emphases on the different error types. Those questions can first be ranked and then used to narrow down the possible options.

We argue that formalizing the selection process in form of a decision tree is the best approach to make this procedure transparent and straightforward. Several promising attempts have been made to provide guidance in this matter~\cite{Saleiro2018,Ruf2020,Makhlouf2020}. In~\cite{Ruf2021} we propose our own process for this purpose. While we acknowledge that designing such a structure is a topic of itself which goes beyond the scope of this position paper, the latest findings in this research field make us confident to expect the arrival of robust and actionable solutions soon. As a prerequisite for the present work, we therefore anticipate those results and assume the existence of a commonly accepted decision tree which formalizes the process of identifying the best fairness definition for a given use case. 

\begin{figure}[h]
  \centering
  \includegraphics[width=1\linewidth]{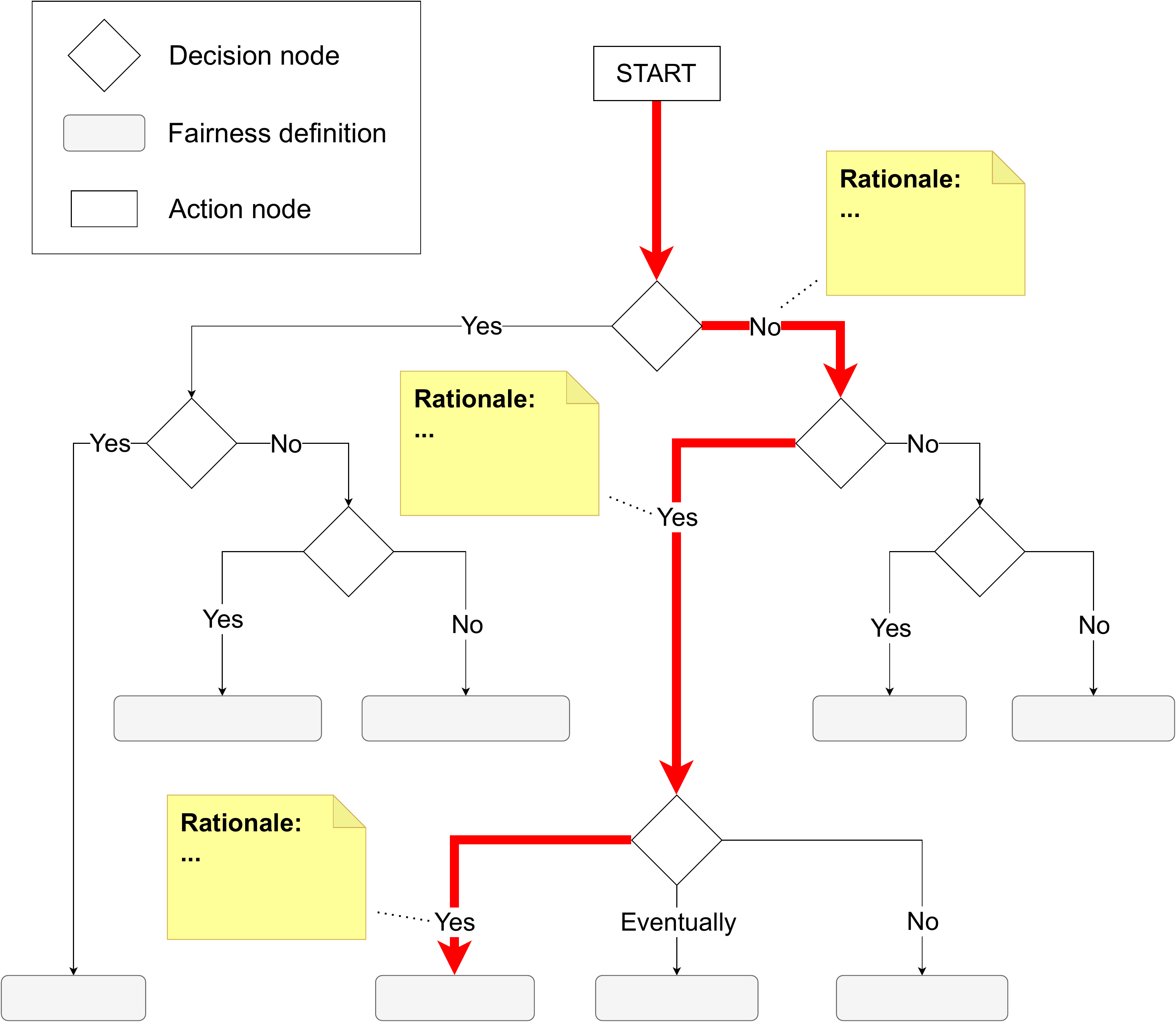}
  \caption{Decision path (red arrows) in mock decision tree with explanations (yellow notes) which explain the choice of fairness definition to the user.}
  \label{fig:mock_tree}
\end{figure}

\section{Explain and Justify Your Choice}
Now the core proposition here is to use such a decision tree not only to identify the best fairness definition, but also as an instrument to back the result and to communicate the reasoning behind the decision. In Figure~\ref{fig:mock_tree}, we illustrate the idea with a mock decision tree. The decision path which leads to a particular fairness definition is highlighted, and each decision is justified with context-specific reasoning. It is worth to note that the final result depends on reasons and ethical principles that are expressed in the decisions – an optimal choice of universal validity does not exist for this problem.

While the problem of bias in AI may be complex and many people lack technical understanding of the field, most of them are very familiar with principal questions around fairness. Instead of requiring the end user to generally trust the AI application because it satisfies some sort of fairness, we propose to engage with him or her on a deeper level: Describe the implemented fairness definition in theory and by example, but also offer insight in the selection process to show how this choice was carefully considered.

\section{Illustration: Rapid Covid-19 Testing}
We illustrate our approach with a sample scenario. The use case is a new, AI-based rapid Covid-19 testing application which is capable of delivering results within a few minutes only, hence much faster than conventional tests, but with lower accuracy. As sensitive subgroups we consider different age groups; for instance people below 60 years of age, and anyone older. 

\begin{figure}[h]
  \centering
  \includegraphics[width=0.8\linewidth]{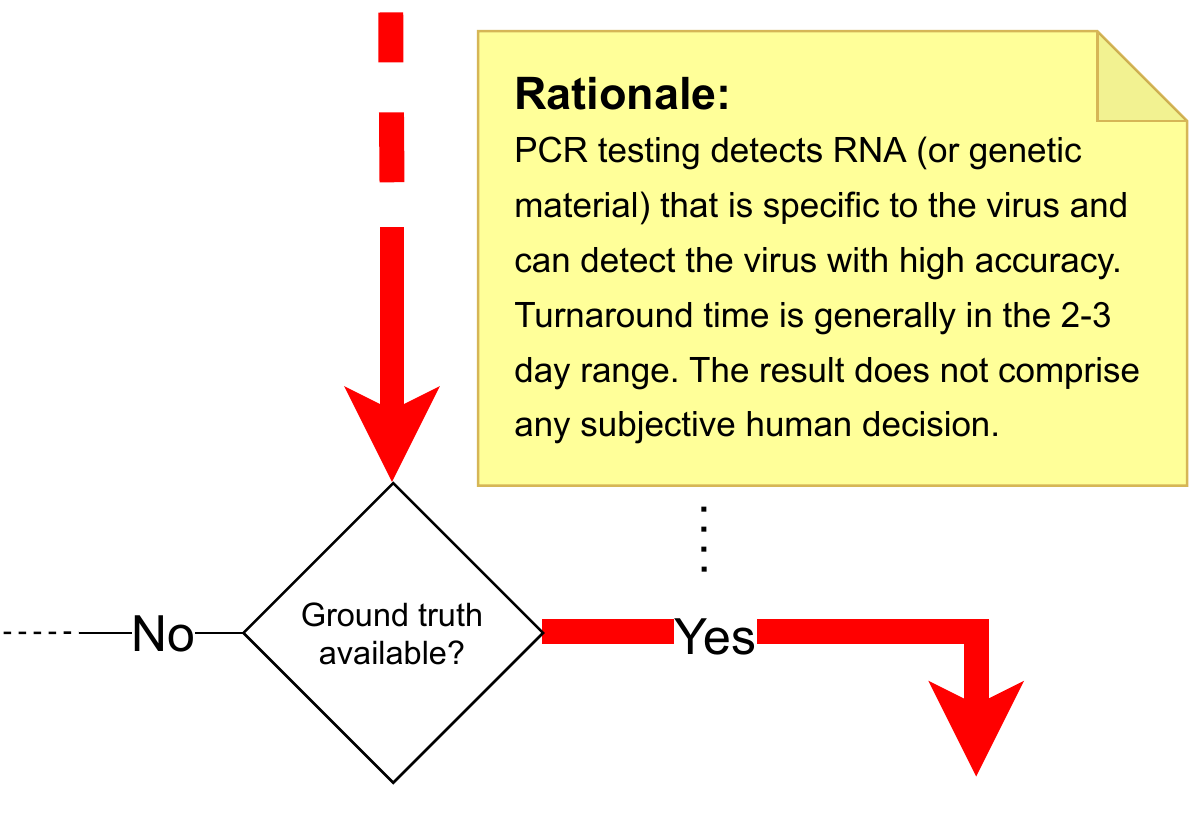}
  \caption{Sample explanation in a medical application for rapid Covid-19  detection. The argument is that the ground truth (here reliable diagnosis results) is available in the given scenario. }
  \label{fig:example}
\end{figure}

In any given decision tree designed to identify the most appropriate fairness definition for this application, one decision node will definitely relate to the availability of  unbiased ground truth data, hence the consideration if the training data are free of any subjective human decisions. Depending on the answer, different fairness definitions become eligible. In the case of rapid Covid-19 testing, the answer would be "yes", because another test (PCR) exists and can detect the virus with high accuracy.
Figure~\ref{fig:example} shows a possible, context-related reasoning for this decision point.\\

This decision would be stored and made, together with all other considerations, available to the public, allowing then to at least explain or even justify how fairness was defined in this context. 

As proof of concept, we elaborated a complete study of the rapid Covid-19 example. 
For the sake of accessibility, we developed an interactive online tool to showcase the result. In this version, info boxes with extended details, examples and references facilitate navigating the tree. The decision path is highlighted in the diagram and the reasoning behind each decision can be displayed as tooltips. The online tool is available for review on our GitHub page\footnote{\url{https://axa-rev-research.github.io/explain-fairness-demo.html}}.


\section{Conclusion}
We propose measures to maintain user confidence in AI, especially with respect to fairness. First, we recommend to formalize the selection process of an adapted fairness definition for any given AI application. In particular, we argue that decision trees are well suited for this task. We further suggest to use the same structure to explain to the end users why the implemented fairness objective was considered to be the most appropriate. With this approach, including our interactive online tool, we hope to provide operational support for implementing human-centered, fair AI in practice. 

\bibliographystyle{SIGCHI-Reference-Format}
\bibliography{sample}

\end{document}